\documentclass{article}

\usepackage[final, nonatbib]{neurips_2025}

\usepackage[utf8]{inputenc} 
\usepackage[T1]{fontenc}    
\usepackage{hyperref}       
\usepackage{url}            
\usepackage{booktabs}       
\usepackage{amsfonts}       
\usepackage{nicefrac}       
\usepackage{microtype}      
\usepackage{xcolor}         
\usepackage{graphicx}
\usepackage{booktabs}
\usepackage{multirow}
\usepackage{subcaption}
\usepackage{array}
\usepackage{colortbl}
\usepackage{xcolor}
\definecolor{vireo}{RGB}{126,3,168}
\usepackage{amsmath}
\usepackage[svgnames, x11names]{xcolor}
\usepackage{capt-of}
\usepackage{wrapfig}
\usepackage{graphicx}
\usepackage{placeins}

\definecolor{bestBase}{HTML}{D44478}
\definecolor{secondBase}{HTML}{FFAA33}
\definecolor{thirdBase}{HTML}{FFDD33}

\colorlet{bestLight}{bestBase!20!white}
\colorlet{secondLight}{secondBase!20!white}
\colorlet{thirdLight}{thirdBase!20!white}

\title{Leveraging Depth and Language for Open-Vocabulary Domain-Generalized Semantic Segmentation}

\author{
Siyu Chen\textsuperscript{\rm 1,2 †} \quad Ting Han\textsuperscript{\rm 2,3 † *} \quad Chengzheng Fu\textsuperscript{\rm 4 ‡} \quad Changshe Zhang\textsuperscript{\rm 5 ‡} \quad Chaolei Wang\textsuperscript{\rm 3 ‡} \\
\quad \textbf{Jinhe Su}\textsuperscript{\rm 1 *} \quad \textbf{Guorong Cai}\textsuperscript{\rm 1}  \quad \textbf{Meiliu Wu}\textsuperscript{\rm 2 *}
    \vspace{5pt} \\
    \textsuperscript{\rm 1} Jimei University, \textsuperscript{\rm 2} University of Glasgow, \textsuperscript{\rm 3} Sun Yat-sen University,\\
    \textsuperscript{\rm 4} Nanjing University of Aeronautics and Astronautics, 
    \textsuperscript{\rm 5} Xidian University,  \\
}

\begin{document}

\maketitle

\begin{abstract}
Open-Vocabulary semantic segmentation (OVSS) and domain generalization in semantic segmentation (DGSS) highlight a subtle complementarity that motivates Open-Vocabulary Domain-Generalized Semantic Segmentation (OV-DGSS). OV-DGSS aims to generate pixel-level masks for unseen categories while maintaining robustness across unseen domains, a critical capability for real-world scenarios such as autonomous driving in adverse conditions. We introduce \textbf{\textcolor{vireo}{Vireo}}, \textbf{a novel single-stage framework for OV-DGSS that unifies the strengths of OVSS and DGSS for the first time.} Vireo builds upon the frozen Visual Foundation Models (VFMs) and incorporates scene geometry via Depth VFMs to extract domain-invariant structural features. To bridge the gap between visual and textual modalities under domain shift, we propose three key components: (1) GeoText Query, which align geometric features with language cues and progressively refine VFM encoder representations; (2) Coarse Mask Prior Embedding (CMPE) for enhancing gradient flow for faster convergence and stronger textual influence; and (3) the Domain-Open-Vocabulary Vector Embedding Head (DOV-VEH), which fuses refined structural and semantic features for robust prediction. Comprehensive evaluation on these components demonstrates the effectiveness of our designs. Our proposed Vireo achieves the \textbf{state-of-the-art performance and surpasses existing methods by a large margin} in both domain generalization and open-vocabulary recognition, offering a unified and scalable solution for robust visual understanding in diverse and dynamic environments. Code is available at \textcolor{vireo}{\href{https://github.com/SY-Ch/Vireo}{https://github.com/SY-Ch/Vireo}}.
\end{abstract}

\section{Introduction}

Open-Vocabulary Domain-Generalized Semantic Segmentation (OV-DGSS) denotes the joint execution of open-vocabulary semantic segmentation (OVSS)\cite{liang2023open,qin2023freeseg,xie2024sed,han2023open} and domain generalization in semantic segmentation (DGSS)\cite{benigmim2024collaborating,wei2024stronger,niemeijer2024generalization,bi2024learning} tasks. It involves training a model that, without access to target-domain samples or annotations for novel categories, can generate pixel-wise segmentation for unseen classes while sustaining high performance across previously unseen domains (such as different cities, lighting environments, or climatic conditions). For instance, when an autonomous vehicle receives a query like “Can I park next to that bollard?”, its perception system must comprehend the linguistic input and accurately segment the referenced object at the pixel level—even under adverse conditions such as dim lighting, rain-streaked lenses, or region-specific visual appearances.

OVSS and DGSS share notable similarities and can both be implemented using either multi-stage or single-stage strategies. In multi-stage OVSS \cite{liang2023open}, candidate regions or coarse masks are generated and then classified by a text approach, while multi-stage DGSS \cite{ISW}first aligns domains through adversarial alignment or style transfer and then trains the segmentation approach on the aligned features. 

In single-stage OVSS \cite{cho2024cat,luo2023segclip}, the segmentation head is dynamically conditioned on text prompts to directly produce masks for each class. In contrast, single-stage DGSS \cite{wu2022siamdoge,peng2022semantic} incorporates domain-invariant modules within the backbone or segmentation head, enabling the model to jointly learn segmentation and generalization in a unified forward pass without requiring distinct stages. The key distinction lies in their focus: OVSS requires integration of visual features with textual semantics to accurately identify unseen classes, whereas DGSS emphasizes robustness to domain shifts. 

Consequently, integrating open-vocabulary recognition of novel classes with domain robustness into a unified framework presents two primary challenges: (1) \textit{\textbf{Text–vision alignment modules often degrade outside the source domain, leading to significant performance drops even for previously seen classes.}} (2) \textit{\textbf{Domain-invariant strategies may suppress fine-grained semantic cues, hindering the model’s ability to precisely respond to detailed textual queries.}}

\begin{figure}
    \centering
    \setlength{\abovecaptionskip}{0.1cm}
    \setlength{\belowcaptionskip}{-0.5cm}
    \includegraphics[width=1\linewidth]{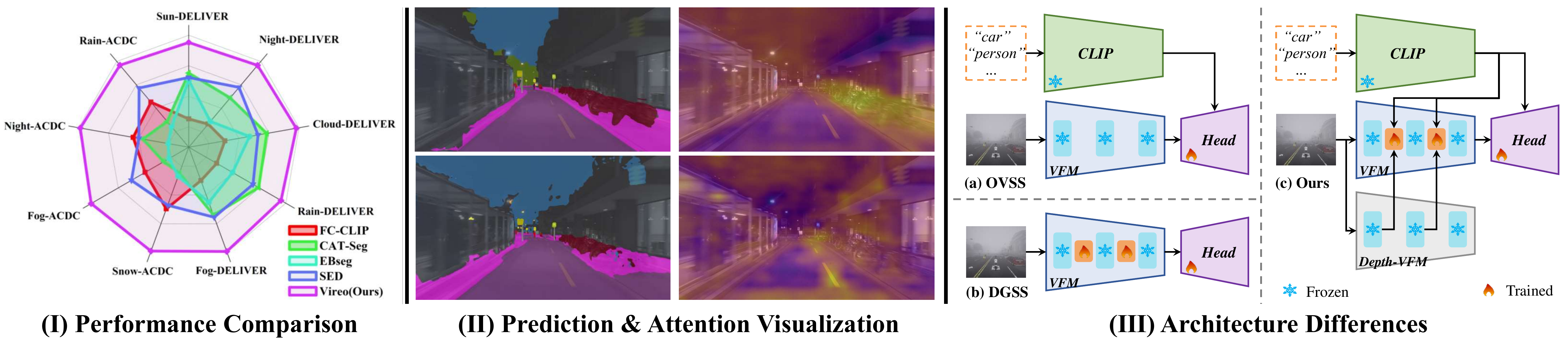}
    \caption{\textbf{Overview of Vireo and effectiveness.} (I) Performance comparison under various adverse conditions across ACDC and DELIVER datasets, showing that Vireo consistently outperforms existing methods. (II) Qualitative visualization of segmentation predictions (left) and attention maps (right) under extreme night scenes, illustrating Vireo’s robustness and precise alignment with semantic cues. (III) Architectural comparison: (a) Traditional OVSS and (b) DGSS pipelines freeze or fine-tune VFM separately without cross-modal integration; (c) Our proposed Vireo introduces GeoText Query and Depth-VFM integration to enhance both semantic alignment and domain robustness.}
    \label{fig: fig1}
\end{figure}

Recent single-stage DGSS studies have increasingly adopted strategies that fine-tune learnable tokens across the layers of a frozen Visual Foundation Model (VFM) \cite{caron2021emerging,oquab2023dinov2,radford2021learning,yang2024depth,yang2024depth_v2} to adapt its feature representations. In OVSS, the VFM encoder is typically fully frozen, with efforts focused on designing the decoder to endow the model with open-vocabulary recognition capabilities. This reveals a subtle complementarity between the two paradigms: \textbf{DGSS emphasizes the encoder by leveraging the VFM’s strong feature generalization to learn cross-domain representations, whereas OVSS freezes the encoder and emphasizes the decoder to enable open-vocabulary recognition.}

Moreover, in cross-domain scenarios, depth and geometric cues are largely insensitive to variations in illumination and texture\cite{chen2025hspformer, han2024epurate}. They supply reliable spatial constraints, easing the distribution shift of RGB features and sharpening boundary localization. Recent studies such as DepthForge \cite{chen2025stronger} have shown that injecting depth query into a frozen VFM boosts domain generalization. Motivated by these findings, we adopt DepthAnything V2 \cite{depth_anything_v2} as our Depth VFM: its diverse pre-training delivers consistent depth estimation across domains, and keeping it fully frozen incurs minimal training cost while preserving real-time inference speed.

In this paper, we propose a VFM-based single-stage pipeline for OV-DGSS, termed \textbf{Vireo}. Specifically, in the encoder, both the VFM and DepthAnything modules are kept frozen. The VFM is leveraged to robustly capture cross-domain visual features, while DepthAnything extracts the scene’s intrinsic geometric structure. On this basis, we introduce \textbf{Geometric-Text Query (GeoText Query)} to fuse the extracted structural features with manually provided textual cues, progressively refining the feature maps between the frozen VFM layers. To mitigate the slow convergence caused by sparse gradients propagating through the frozen encoder, we introduce a \textbf{Coarse Mask Prior Embedding (CMPE)} at the beginning of the decoder to inject denser gradient signals. This design accelerates the convergence of mask supervision and further strengthens the influence of textual priors. Subsequently, we design the \textbf{Domain-Open-Vocabulary Vector Embedding Head (DOV-VEH)} to strengthen the synergistic integration of structural and textual modalities, ensuring that both cross-domain structural features and open-vocabulary cues learned by the GeoText Query are fully utilized in the final prediction.

\textbf{For challenge 1}: We find that GeoText Query not only capture structural and semantic cues within the frozen VFM encoder but also guide the progressive refinement of its feature representations. \textbf{For challenge 2}: The CMPE enhances gradient back-propagation into the encoder, while the redesigned DOV-VEH deepens the fusion of visual and textual priors. These components form a unified framework that simultaneously achieves domain robustness and strong open-vocabulary generalization. Our main contributions are summarized as follows:

(1) We propose \textbf{Vireo}, a novel VFM-based single-stage framework for OV-DGSS.

(2) We introduce \textbf{GeoText Query} to progressively refine frozen VFM features by injecting geometric cues from DepthAnything and aligning them with textual semantics, enabling structural-semantic fusion across encoder layers.

(3) We design two complementary modules—\textbf{CMPE} for enhancing gradient flow and \textbf{DOV-VEH} for fusing visual and textual priors—together boosting segmentation performance under domain shifts and unseen classes.

\begin{figure}[!t]
  \centering
  \setlength{\abovecaptionskip}{0.1cm}
  \setlength{\belowcaptionskip}{-0.5cm}
    \includegraphics[width=1\linewidth]{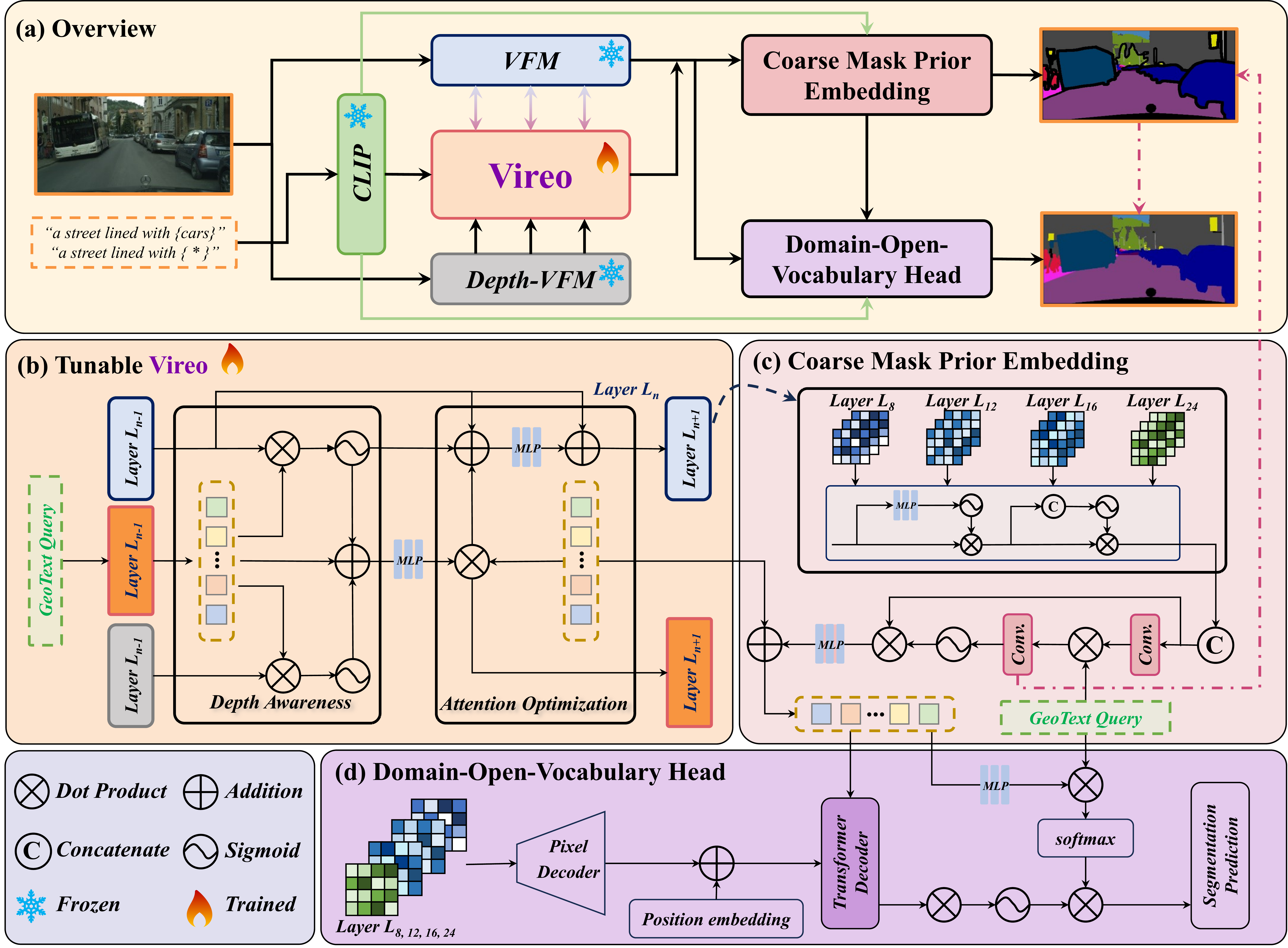}%
  \caption{\textbf{Overview of the proposed \textcolor{vireo}{Vireo} framework for OV-DGSS.} (a) \textbf{Framework Overview.} Geometric-Text Query permeate our model: they’re injected in Tunable Vireo to align domain priors, reused in CMPE to guide multi-scale feature fusion, and employed in DOV-VEH as queries for open-vocabulary segmentation, forming a unified end-to-end loop. (b) \textbf{Tunable Vireo.} GeoText Query are used to inject structural-textual priors across multiple layers. Depth-aware fusion and attention optimization are applied between intermediate frozen layers  to progressively refine representations. (c) \textbf{Coarse Mask Prior Embedding (CMPE).} Multi-scale features from VFM are combined with GeoText Query to provide dense supervision and gradient signals for downstream modules. (d) \textbf{Domain-Open-Vocabulary Head (DOV-VEH).} Multi-level features are processed by a Pixel Decoder and Transformer Decoder to produce final predictions guided by open-vocabulary text queries. }
  \label{fig: architecture}
\end{figure}

\section{Related Works}

\textbf{Open-Vocabulary Semantic Segmentation}. Open-Vocabulary Semantic Segmentation (OVSS) aims to segment objects based on arbitrary textual descriptions, moving beyond fixed, pre-defined categories. A key enabler for OVSS \cite{li2022language,zhou2022extract,liang2023open,xu2023open} has been vision-language models (VLMs), particularly CLIP \cite{radford2021learning}, which aligns visual and textual representations. 

\textbf{Domain-Generalized Semantic Segmentation}. Domain Generalization in Semantic Segmentation (DGSS) addresses the performance degradation when models encounter unseen target domains due to domain shift—variations in data distributions (e.g., lighting, weather). Data augmentation \cite{hoffman2018cycada, volpi2018generalizing} and learning domain-invariant representations \cite{ganin2016domain,choi2021robustnet} are two key strategies to enhance model robustness.
This detailed discussion in the appendix further contextualizes our specific contributions within the broader landscape of these research areas.

\section{Methodology - Vireo}

We formulate the open-vocabulary DGSS problem as predicting a pixel-wise segmentation mask from an input image and a set of text labels. Given an image  $I$ and a set of open-vocabulary class names $C$, our goal is to produce a mask $M$ where each pixel is classified according to one of the classes in $C$. As illustrated in Figure~\ref{fig: architecture}, our framework Vireo tackles this by leveraging frozen vision and text encoders with additional trainable prompt and depth modules for improved generalization, it consists of three key modules tailored for the OV-DGSS: (1) \textbf{Tunable Vireo with GeoText Query:} Introduces GeoText Query to inject and refine both geometric and textual information across layers of a frozen VFM. (2) \textbf{Coarse Mask Prior Embedding (CMPE):} Generates coarse prior masks to guide segmentation and reinforce gradient flow from the decoder back to the frozen encoder. (3) \textbf{Domain-Open-Vocabulary Vector Embedding Head (DOV-VEH):} Integrates visual, geometric, and semantic features to produce final OV-DGSS predictions.

\subsection{Overview}

The input RGB image $I \in \mathbb{R}^{H \times W \times 3}$ is duplicated and independently fed into two frozen encoders: a VFM encoder $\mathcal{F}^{V}$ and a DepthAnything encoder $\mathcal{F}^{D}$. These backbones extract a series of multi-scale features $f^{V}_{l}$ and $f^{D}_{l}$ respectively, where $l$ denotes the encoder layer index in $[1, L]$. In parallel, a set of class labels $\mathcal{C} = \{c_{1}, \dots, c_{K}\}$ is transformed into natural language prompts $T = \{\texttt{prompt}_{1}, \dots, \texttt{prompt}_{K}\}$, which are encoded using a frozen CLIP text encoder $\mathcal{F}^{T}$ to obtain textual embeddings $t_{k}\in \mathbb{R}^d$. These features serve as shared semantic priors across the entire framework and are precomputed once at initialization. In the Tunable Vireo module, each layer receives a tuple $(f^{V}_{l}, f^{D}_{l}, t_{k})$, and applies the GeoText Query $ P_{l} $ to align and refine the visual features. This is achieved by computing cross-modal attention maps $\mathcal{A}_l = \text{Attn}(P_l, f^{V}_{l}, f^{D}_{l}, t_{k})$ followed by feature fusion and projection. The refined output $\hat{f^{V}}_{l}$ is then forwarded back to update the VFM’s layer-wise activations.

We select the refined visual features $\{{f}^{V}_{l_i}\}_{i=1}^4 $ from the VFM encoder at layers $l_1 = 8$, $l_2 = 12$, $l_3 = 16$, and $l_4 = 24$ for downstream decoding. In the Coarse Mask Prior Embedding (CMPE) module, these features are first upsampled to a unified spatial resolution and passed through a channel-spatial attention gating function $\mathcal{G}(\cdot)$. The gated outputs are then fused to form a global coarse feature $ f^{M} \in \mathbb{R}^{d} $. The $ f^{M} $ is projected and matched with the text embeddings $t_k \in \mathbb{R}^{d}$ to produce a coarse class probability map $ \mathcal{M} \in \mathbb{R}^{H \times W \times K} $, where $\mathcal{M} $ serves both as a weak supervision signal and as a prior to construct the query embeddings $ \mathcal{Q} $. The query prior is added to the GeoText Query and forwarded to the final segmentation head. In the Domain-Open-Vocabulary Vector Embedding Head (DOV-VEH), the multi-scale features $\{\hat{f}^{V}_{l_i}\}_{i=1}^4$ are passed through a pixel decoder $\mathcal{D}_{p}(\cdot)$ to enhance spatial representation, followed by a Transformer decoder $\mathcal{D}_{T}(\cdot)$ that leverages positional embedding. The GeoText Query serve as learnable queries and interact with both the decoded features and the text embeddings, producing pixel-level mask embeddings $\mathcal{E}_{\text{mask}}(x,y) \in \mathbb{R}^{d}$ and classification embeddings $\mathcal{E}_{\text{cls}}(k) \in \mathbb{R}^{d}$. The final prediction is $ \hat{\mathcal{M}}(x, y, k) \in \mathbb{R}^{H \times W \times K}$ provides the pixel-wise semantic prediction with both fine-grained detail and open-vocabulary generalization capability.

\subsection{Tunable Vireo with GeoText Query}

To improve efficiency, we first precompute and share the textual prompt embeddings using a frozen CLIP text encoder $\mathcal{F}^T$. Specifically, a set of class labels $\mathcal{C} = \{c_1, \dots, c_K\}$ is transformed into language prompts $T = \{\texttt{prompt}_1, \dots, \texttt{prompt}_K\}$, which are encoded as $t_k = \mathcal{F}^T(\texttt{prompt}_k)$, where $t_k \in \mathbb{R}^d$. These embeddings are reused across all GeoText Prompt layers, CMPE, and DOV-VEH modules to avoid redundant computation.

During inference, the input image $I \in \mathbb{R}^{H \times W \times 3}$ is simultaneously processed by a frozen visual encoder $\mathcal{F}^V$ and a frozen depth encoder $\mathcal{F}^D$ (e.g., DepthAnything). For each selected layer $l \in \{1, \dots, L\}$, we obtain the visual feature map $f^V_l = \mathcal{F}^V_l(I)$ and the depth feature map $f^D_l = \mathcal{F}^D_l(I)$. Each layer of the Tunable Vireo module receives the tuple $(f^V_l, f^D_l, \{t_k\})$ along with a layer-specific GeoText Prompt $P_l \in \mathbb{R}^{N \times d}$, where $N$ is the number of learnable queries.

The prompt $P_l$ first interacts with the textual embeddings via a fusion block, then attends to both $f^V_l$ and $f^D_l$ through cross-attention mechanisms:
\begin{equation}
\mathcal{A}_l = \text{CrossAttn}(P_l,\ f^V_l,\ f^D_l,\ \{t_k\}).
\end{equation}

The attention outputs are fused via weighted summation, then passed through an MLP projection layer and multiplied element-wise with $P_l$. A residual connection adds this result to the original feature map $f^{V}_{l}$, yielding a refined visual representation $\hat{f}^{V}_{l}$. Finally, another MLP transforms $\hat{f}^{V}_{l}$ into the input for the next VFM layer, and the updated GeoText Prompt $P_{l+1}$ is passed forward.

This progressive refinement continues across all selected layers, enabling the model to inject and align geometric and semantic information at multiple scales, thereby enhancing cross-domain robustness and open-vocabulary generalization.

\subsection{Coarse Mask Prior Embedding (CMPE)}

We select the refined visual features $\{\hat{f}^{V}_{l_i}\}_{i=1}^4 $ from the VFM encoder at layers $l_1 = 8$, $l_2 = 12$, $l_3 = 16$, and $l_4 = 24$, respectively. Each feature map is upsampled to a common spatial resolution $(H \times W)$ via bilinear interpolation, and then passed through an Adaptive Attention Gate (AAG) $\mathcal{G}(\cdot)$, which enhances informative channels and spatial regions. Specifically, AAG applies two $1 \times 1$ convolutions followed by ReLU and Sigmoid for channel attention, and a $3 \times 3$ convolution followed by Sigmoid for spatial attention.

The attended features are concatenated along the channel axis and fused via a $1 \times 1$ convolution to restore the embedding dimension $d$, yielding a fused feature representation: $f^{M} = \text{Fuse}( \mathcal{G}(\hat{f}^{V}_{l_i}))$. We apply a residual addition between $f^{M}$ and the final layer output $\hat{f}^V_{l_4}$ to obtain the updated mask feature: $ f^{M} = f^{M} + \hat{f}^{V}_{l_4}$. This fused feature $f^{M}(x, y) \in \mathbb{R}^d$ is projected to the same dimension as the text embeddings $t_k \in \mathbb{R}^d$ and compared via Einstein summation to generate a coarse semantic probability map $\mathcal{M}(x, y, k) = \left\langle f^{M}(x,y),\ t_k \right\rangle$ , where $\mathcal{M} \in \mathbb{R}^{B \times K \times H \times W}$. This coarse mask is supervised with a segmentation loss to enhance gradient flow through the frozen encoder.

To generate query priors for the downstream segmentation head, we first normalize $\mathcal{M}$ across the spatial domain to derive attention weights:
\begin{equation}
\alpha_{k}(x,y) = \frac{\exp(\mathcal{M}(x,y,k))}{\sum_{x',y'} \exp(\mathcal{M}(x',y',k))}.
\end{equation}
Then, we compute the class-specific aggregated feature by spatially weighting $f^{M}$:
\begin{equation}
f^{\text{class}}_k = \sum_{x,y} \alpha_k(x,y) \cdot f^{M}(x,y).
\end{equation}
Each $f^{\text{class}}_k$ is projected into the embedding space as $e^{\text{class}}_k \in \mathbb{R}^d$, and combined with a set of learnable queries vectors $\{q_j\}_{j=1}^{N_q}$ to produce the query priors:
\begin{equation}
q^{\text{prior}}_j = \sum_{k=1}^{K} \text{Softmax}(\langle q_j,\ e^{\text{class}}_k \rangle) \cdot e^{\text{class}}_k.
\end{equation}
The final query priors $\{q^{\text{prior}}_j\}_{j=1}^{N_q}$ are added to the corresponding GeoText Query and forwarded into the Domain-Open-Vocabulary Vector Embedding Head (DOV-VEH).

\subsection{Domain-Open-Vocabulary Vector Embedding Head (DOV-VEH)}

The DOV-VEH module receives the refined multi-scale features $\{\hat{f}^{V}_{l_i}\}_{i=1}^4 $ from the VFM encoder, where $l_1=8$, $l_2=12$, $l_3=16$, and $l_4=24$, along with the updated GeoText Query $\{P_l\}$. These features are first processed by a pixel decoder $\mathcal{D}_{p}(\cdot)$, which leverages multi-scale cross-attention to extract rich spatial context: $f^{\text{pix}} =\mathcal{D}_{p}\left(\hat{f}^{V}_{l_i}\right)$.
The fused feature $f^{\text{pix}} \in \mathbb{R}^{H \times W \times d}$ is then compressed via a $1 \times 1$ convolution and enriched with sinusoidal positional encoding to preserve spatial structure.

The enhanced features are fed into a Transformer Decoder $\mathcal{D}_{T}(\cdot)$, where the GeoText Query act as learnable queries. Through stacked layers of self-attention and cross-attention with $f^{\text{pix}}$, the model captures fine-grained visual-semantic alignment, yielding a set of high-resolution mask features $\mathcal{E}_{\text{mask}}(x, y) \in \mathbb{R}^{d}$ at each spatial position.

Simultaneously, the GeoText Query $\{P\}$ are passed through a two-layer MLP and then interact with the text embeddings $\{t_k\}$ (precomputed from the CLIP text encoder) to produce classification-level representations $\mathcal{E}_{\text{cls}}(k) \in \mathbb{R}^d$. The final segmentation prediction $\hat{\mathcal{M}} \in \mathbb{R}^{H \times W \times K}$ is generated via an Einstein summation over the two embeddings:
\begin{equation}
\hat{\mathcal{M}}(x, y, k) = \sum_{d=1}^{D} \mathcal{E}_{\text{mask}}(x, y, d) \cdot \mathcal{E}_{\text{cls}}(k, d),
\end{equation}
where $D$ is the feature embedding dimension. This design enables DOV-VEH to generate pixel-level segmentation masks that are both spatially accurate and semantically aligned with open-vocabulary textual queries.

\begin{table}[!t]
  \centering
  \setlength{\belowcaptionskip}{-0.5cm}
  \caption{Performance comparison between our Vireo and existing OVSS and DGSS methods under \textit{Citys. $\rightarrow$ ACDC + BDD. + Map.}, and \textit{GTA5. $\rightarrow$ Citys. + BDD. + Map.} generalization settings. Top three results are highlighted as \colorbox{bestLight}{\textbf{best}}, \colorbox{secondLight}{second}, and \colorbox{thirdLight}{third}, respectively. (\%)}
  \label{tab:adaptation}
  \resizebox{\textwidth}{!}{%
  \begin{tabular}{c c *{7}{c} *{3}{c}}
    \toprule
    \multirow{2}{*}{Method}
      & \multirow{2}{*}{Proc.\ \& Year}
      & \multicolumn{7}{c}{Trained on Cityscapes}
      & \multicolumn{3}{c}{Trained on GTA5}\\
    \cmidrule(lr){3-9} \cmidrule(lr){10-12}
    & 
      & Night-ACDC & Fog-ACDC & Rain-ACDC & Snow-ACDC
      & BDD 100k & Mapillary & GTA5
      & Cityscapes & BDD100k & Mapillary
       \\
      \midrule
      \multicolumn{12}{c}{\textbf{OVSS Method}} \\
    FC-CLIP~\cite{yu2023convolutions} 
      & NeurIPS2023
      & 40.8 & 64.4 & 63.2  & 61.5
      & 55.92 & 66.12 & 47.12
      & 53.54 & 51.41 & 58.60 \\
    EBSeg~\cite{shan2024open} 
      & CVPR2024
      & 27.7 & 56.5 & 51.8 & 50.1
      & 48.91 & 63.40 & 42.61
      & 44.80 & 40.59 & 56.28 \\
    CAT-Seg~\cite{cho2024cat}
      & CVPR2024
      & 37.2 & 58.3 & 45.6 & 49.0
      & 48.26 & 54.74 & 45.18
      & 43.52 & 44.28 & 50.88 \\
    SED~\cite{xie2024sed}
      & CVPR2024
      & 38.7 & 69.0 & 56.4  & 60.2
      & 53.30 & 64.32 & 48.93
      & 47.45 & 48.16 & 57.38 \\
      \midrule
    \multicolumn{12}{c}{\textbf{DGSS Method}} \\
    \multicolumn{2}{l}{\textit{ResNet based:}} \\
    IBN~\cite{IBN} 
      & ECCV2018 
      & 21.2 & 63.8 & 50.4  & 49.6
      & 48.56 & 57.04 & 45.06 
      & - & - & - \\
    RobustNet~\cite{ISW}
      & CVPR2021
      & 24.3 & 64.3 & 56.0  & 49.8
      & 50.73 & 58.64 & 45.00
      & 36.58 & 35.20 & 40.33 \\
    WildNet~\cite{WildNet} 
      & CVPR2022
      & 12.7 & 41.2 & 34.2  & 28.4
      & 50.94 & 58.79 & 47.01
      & 44.62 & 38.42 & 46.09 \\
     \multicolumn{2}{l}{\textit{Transformer based:}}  \\
    HGFormer~\cite{HGFormer} 
      & CVPR2023
      & 52.7 & 69.9 & 72.0  & 68.6
      & 53.40 & 66.90 & 51.30 
      & - & - & - \\
    CMFormer~\cite{CMFormer} 
     & AAAI2024
      & 33.7 & 77.8 & 67.6  & 64.3
      & 59.27 & 71.10 & 58.11 
      & 55.31 & 49.91 & 60.09 \\
      \multicolumn{2}{l}{\textit{VFM based:}}\\
    REIN~\cite{REIN} 
      & CVPR2024
      & \cellcolor{thirdLight}\textbf{55.9}
      & \cellcolor{thirdLight}\textbf{79.5}
      & \cellcolor{thirdLight}\textbf{72.5}
      & \cellcolor{thirdLight}\textbf{70.6}
      & \cellcolor{thirdLight}\textbf{63.54}
      & \cellcolor{thirdLight}\textbf{74.03}
      & \cellcolor{thirdLight}\textbf{62.41}
      & \cellcolor{thirdLight}\textbf{66.40}
      & \cellcolor{thirdLight}\textbf{60.40}
      & \cellcolor{thirdLight}\textbf{66.10} \\
    FADA~\cite{FADA} 
      & NeurIPS2024
      & \cellcolor{secondLight}\textbf{57.4}
      & \cellcolor{secondLight}\textbf{80.2}
      & \cellcolor{secondLight}\textbf{75.0}
      & \cellcolor{secondLight}\textbf{73.5}
      & \cellcolor{secondLight}\textbf{65.12}
      & \cellcolor{secondLight}\textbf{75.86}
      & \cellcolor{secondLight}\textbf{63.78}
      & \cellcolor{secondLight}\textbf{68.23}
      & \cellcolor{secondLight}\textbf{61.94}
      & \cellcolor{secondLight}\textbf{68.09} \\
      \midrule
      \multicolumn{12}{c}{\textbf{OV-DGSS Method}} \\
    \textbf{\textcolor{vireo}{Vireo (Ours)}} 
      & -
      & \cellcolor{bestLight}\textbf{60.6}
      & \cellcolor{bestLight}\textbf{82.3}
      & \cellcolor{bestLight}\textbf{76.3}
      & \cellcolor{bestLight}\textbf{76.2}
      & \cellcolor{bestLight}\textbf{66.73}
      & \cellcolor{bestLight}\textbf{75.99}
      & \cellcolor{bestLight}\textbf{67.86}
      & \cellcolor{bestLight}\textbf{70.69}
      & \cellcolor{bestLight}\textbf{62.91}
      & \cellcolor{bestLight}\textbf{69.63} \\
    \bottomrule
  \end{tabular}
  }
\end{table}

\section{Experiments}

\vspace{-2mm}

\subsection{Datasets \& Evaluation Protocols}

We evaluate Vireo on six real-world datasets (\textbf{Cityscapes} \cite{cityscapes}, \textbf{BDD100K} \cite{bdd100k}, \textbf{Mapillary} \cite{mapillary}, \textbf{ACDC} \cite{9711067}, \textbf{ADE150} \cite{zhou2019semantic}, and \textbf{ADE847} \cite{zhou2019semantic}) and two synthetic datasets (\textbf{GTA5} \cite{gtav} and \textbf{DELIVER} \cite{zhang2023delivering}). \textbf{Cityscapes} (City.) is an autonomous-driving dataset with 2,975 training images and 500 validation images, each at a resolution of $2048\times1024$. \textbf{BDD100K} (BDD.) and \textbf{Mapillary} (Map.) provide 1,000 and 2,000 validation images, respectively, at resolutions of $1280\times720$ and $1920\times1080$. \textbf{ACDC} offers 406 validation images captured under extreme conditions (night, snow, fog, and rain), each at $1920\times1080$. \textbf{GTA5} is a synthetic dataset that contains 24,966 labeled images obtained from a video game. \textbf{DELIVER} is a multimodal synthetic dataset comprising 3,983 training images, 2,005 validation images, and 1,897 test images across five weather conditions (cloudy, foggy, night, rainy, and sunny); every image is $1042\times1042$ and there are 25 classes. \textbf{ADE150} and \textbf{ADE847} refer to subsets of the ADE20K dataset \cite{zhou2019semantic}, each containing 2,000 validation images of variable resolution sourced from diverse scenes such as SUN and Places, covering 150 and 847 semantic categories, respectively.

Following the existing DGSS evaluation protocol, we train on one dataset as the source domain and validate on multiple unseen target domains. The three standard evaluation setups are: (1) Cityscapes $\rightarrow$ ACDC; (2) GTA5 $\rightarrow$ Cityscapes, BDD100K, Mapillary; (3) Cityscapes $\rightarrow$ BDD100K, Mapillary, GTA5. To assess our proposed OV-DGSS approaches and compare its open-vocabulary capability against OVSS approaches, we additionally introduce two more configurations: (4) Cityscapes $\rightarrow$ DELIVER, ADE150, ADE847; (5) GTA5 $\rightarrow$ DELIVER, ADE150, ADE847. The evaluation metric is mean Intersection over Union (mIoU).

\subsection{Deployment Details \& Parameter Settings}

Our implementation is built upon the MMSegmentation \cite{mmseg2020} codebase. We employ the AdamW optimizer with an initial learning rate of 1e-4, a weight decay of 0.05, epsilon set to 1e-8, and beta parameters of (0.9, 0.999). The total number of training iterations is 40,000, matching REIN, and we adopt a polynomial learning-rate decay schedule that reduces the learning rate to zero over 40,000 iterations with a decay power of 0.9 and no epoch-based warmup. Data augmentation comprises multi-scale resizing, random cropping (with fixed crop size and category-ratio constraint), random horizontal flipping, and photometric distortion. All experiments are conducted on an NVIDIA RTX A6000 GPU with a batch size of 8, taking approximately 14 hours to train and peaking at around 45 GB of GPU memory usage.

\subsection{Performance Comparison}

\begin{figure}
    \centering
    \setlength{\abovecaptionskip}{0.1cm}
    \setlength{\belowcaptionskip}{-0.5cm}
    \includegraphics[width=1\linewidth]{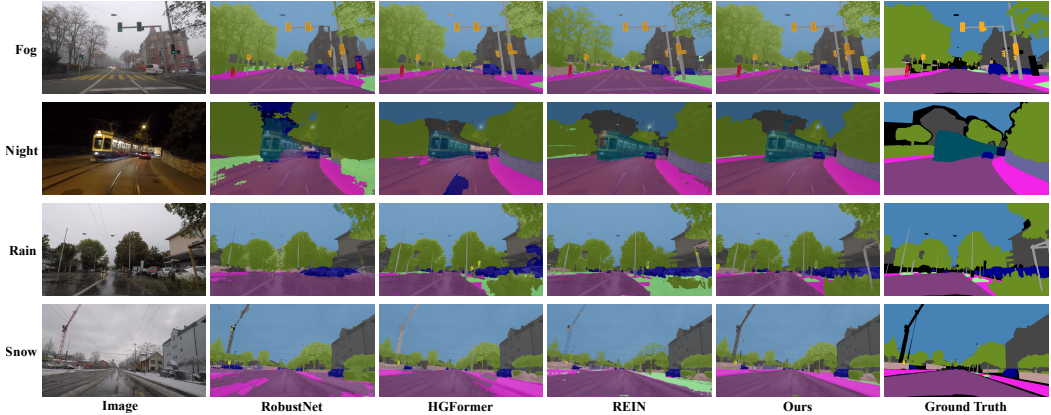}
    \caption{Key segmentation examples of existing DGSS methods and Vireo under the Citys. $\rightarrow$ ACDC unseen target domains under Night, Snow, Rain, and Fog conditions.}
    \label{fig:DGSS_results}
\end{figure}

\begin{table}[!t]
  \centering
  \setlength{\abovecaptionskip}{0.1cm}
  \caption{Performance comparison between our Vireo and existing OVSS methods under \textit{Citys. $\rightarrow$ DELIVER + ADE150 + ADE847} generalization setting. Top three results are highlighted as \colorbox{bestLight}{\textbf{best}}, \colorbox{secondLight}{second}, and \colorbox{thirdLight}{third}, respectively. (\%)}
  \label{tab:deliver-ade}
  \resizebox{\textwidth}{!}{%
  \begin{tabular}{c c *{7}{c}}
    \toprule
    Method
      & Proc.\ \& Year
      & Sun-DELIVER
      & Night-DELIVER
      & Cloud-DELIVER
      & Rain-DELIVER
      & Fog-DELIVER
      & ADE150
      & ADE847 \\
    \midrule
    \multicolumn{9}{c}{\textbf{Train on Cityscapes}} \\
    \multicolumn{2}{l}{\textit{OVSS Method:}} \\
    FC-CLIP~\cite{yu2023convolutions} & NeurIPS2023 
      & 16.93 
      & 14.93 
      & 17.50 
      & 16.59 
      & 17.26 
      & 16.12 
      & \cellcolor{thirdLight}{6.29} \\
    EBSeg~\cite{shan2024open}   & CVPR2024 
      & 26.41 
      & 15.50 
      & 22.62 
      & 20.35 
      & 22.00 
      & 12.75 
      & 3.75 \\
    CAT-Seg~\cite{cho2024cat} & CVPR2024 
      & \cellcolor{secondLight}{28.21}
      & \cellcolor{thirdLight}{20.56}
      & \cellcolor{secondLight}{26.22}
      & \cellcolor{secondLight}{26.53}
      & \cellcolor{thirdLight}{24.80}
      & \cellcolor{secondLight}{20.19}
      & \cellcolor{secondLight}{6.95} \\
    SED~\cite{xie2024sed}     & CVPR2024 
      & \cellcolor{thirdLight}{27.14}
      & \cellcolor{secondLight}{22.79}
      & \cellcolor{thirdLight}{24.40}
      & \cellcolor{thirdLight}{25.18}
      & \cellcolor{secondLight}{25.25}
      & \cellcolor{thirdLight}{18.86}
      & 5.45 \\
      \multicolumn{2}{l}{\textit{OV-DGSS Method:}}\\
    \textbf{\textcolor{vireo}{Vireo (Ours)}} & – 
      & \cellcolor{bestLight}\textbf{35.73}
      & \cellcolor{bestLight}\textbf{27.51}
      & \cellcolor{bestLight}\textbf{32.34}
      & \cellcolor{bestLight}\textbf{31.80}
      & \cellcolor{bestLight}\textbf{32.72}
      & \cellcolor{bestLight}\textbf{21.37}
      & \cellcolor{bestLight}\textbf{7.31} \\
    \midrule
    \multicolumn{9}{c}{\textbf{Train on GTA}} \\
    \multicolumn{2}{l}{\textit{OVSS Method:}} \\
    FC-CLIP~\cite{yu2023convolutions} & NeurIPS2023 
      & 22.24 
      & 18.58 
      & 18.50 
      & 16.59 
      & 19.12 
      & 15.47 
      & 5.73 \\
    EBSeg~\cite{shan2024open}   & CVPR2024 
      & \cellcolor{secondLight}{32.32}
      & 20.05 
      & \cellcolor{thirdLight}{26.19}
      & \cellcolor{thirdLight}{26.19}
      & \cellcolor{secondLight}{28.69}
      & 11.87 
      & 4.19 \\
    CAT-Seg~\cite{cho2024cat} & CVPR2024 
      & \cellcolor{thirdLight}{28.59}
      & \cellcolor{secondLight}{23.49}
      & \cellcolor{secondLight}{27.31}
      & \cellcolor{secondLight}{27.94}
      & \cellcolor{thirdLight}{27.66}
      & \cellcolor{secondLight}{20.45}
      & \cellcolor{secondLight}{7.18} \\
    SED~\cite{xie2024sed}     & CVPR2024 
      & 26.56 
      & \cellcolor{thirdLight}{21.18}
      & 24.95 
      & 24.58 
      & 26.17 
      & \cellcolor{thirdLight}{19.57}
      & \cellcolor{thirdLight}{6.80} \\
    \multicolumn{2}{l}{\textit{OV-DGSS Method:}}\\
    \textbf{\textcolor{vireo}{Vireo (Ours)}} & – 
      & \cellcolor{bestLight}\textbf{38.49}
      & \cellcolor{bestLight}\textbf{29.89}
      & \cellcolor{bestLight}\textbf{33.89}
      & \cellcolor{bestLight}\textbf{33.46}
      & \cellcolor{bestLight}\textbf{35.80}
      & \cellcolor{bestLight}\textbf{21.23}
      & \cellcolor{bestLight}\textbf{7.68} \\
    \bottomrule
  \end{tabular}
  }
\end{table}

\textbf{Domain Generalization Ability}: Table~\ref{tab:adaptation}  summarizes the evaluation results of various state-of-the-art open-vocabulary semantic segmentation (OVSS) and domain-generalized semantic segmentation (DGSS) methods under two cross-domain settings (Cityscapes $\rightarrow$ ACDC, BDD100k, Mapillary, GTA5 and GTA5 $\rightarrow$ Cityscapes, BDD100k, Mapillary). The results demonstrate that our approach achieves outstanding performance across all target datasets, significantly outperforming other OVSS/DGSS methods. Furthermore, the visualizations in Figure~\ref{fig:DGSS_results} shows that Vireo delivers satisfactory predictions in both extreme weather conditions and scenes with dense pedestrian and vehicular traffic.

\begin{figure}[!t]
    \centering
    \includegraphics[width=1\linewidth]{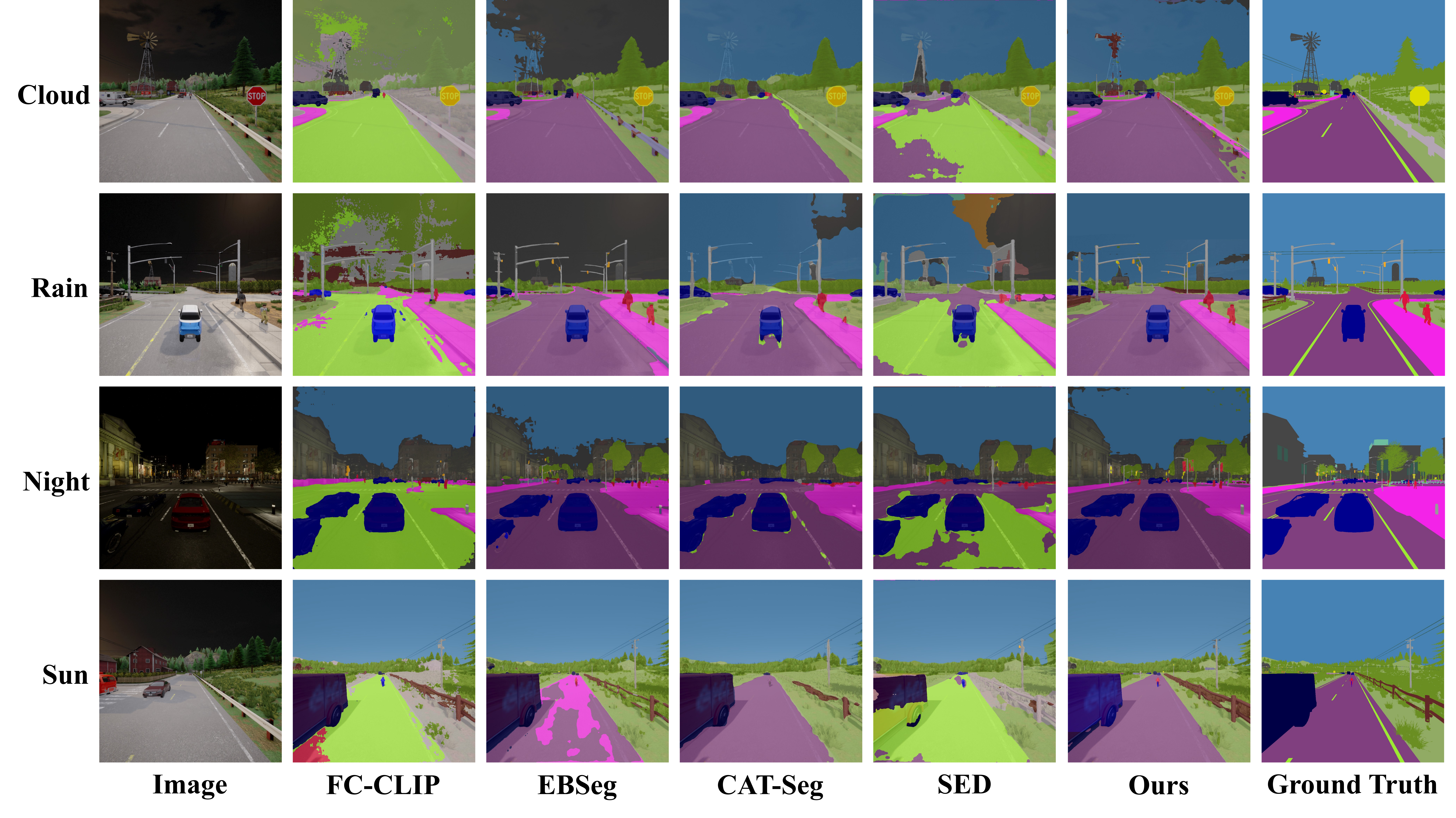}
    \caption{Key segmentation examples of existing OV-DGSS methods comparison on unseen classes and cross domain on Cityscapes → DELIVER.}
    \label{fig:OV-DGSS_result}
\end{figure}

\textbf{Open-Vocabulary Capability}: Table~\ref{tab:deliver-ade} presents a comparison between Vireo and other open-vocabulary semantic segmentation (OVSS) methods under the Cityscapes → DELIVER (sun, rain, night, cloud, fog), ADE150, and ADE847 configurations. The results indicate that conventional OVSS approaches suffer a sharp performance drop in extreme scenarios (e.g., night), whereas our model—enhanced by depth-based geometric features—maintains robust performance, outperforming the strongest OVSS baseline by at least 5\%. Furthermore, as shown in Figure~\ref{fig:OV-DGSS_result}, the coexistence of new open classes and extreme weather conditions in the DELIVER dataset leads OVSS approaches to exhibit significant false positives and false negatives.

Table~\ref{tab:seen} compares Vireo trained on Cityscapes with other open-vocabulary semantic-segmentation methods on the DELIVER dataset across five weather conditions for both Seen and Unseen categories. Vireo achieves the highest mIoU for both groups: it outperforms the second-best method by roughly 7–10 percentage points on Seen classes and by 2–3 points on Unseen ones. Although all methods register lower mIoU on Unseen categories—underscoring the difficulty of open-vocabulary segmentation—Vireo substantially reduces this gap, confirming that its depth-geometry guidance and cross-domain alignment improve recognition of new classes. Overall, Vireo remains consistently superior across weather scenarios and category types, demonstrating stronger out-of-domain generalization.

\begin{table}[!t]
\centering
\caption{Comparison of Seen and Unseen Category mIoU Across Weather Scenarios and Methods.}
\scriptsize
\setlength{\tabcolsep}{2pt}
\renewcommand{\arraystretch}{1.2}
\resizebox{\textwidth}{!}{%
\begin{tabular}{l|*{5}{c}|*{5}{c}|*{5}{c}|*{5}{c}|*{5}{c}}
\toprule
\textbf{Class}
  & \multicolumn{5}{c|}{\textbf{Cloud}}
  & \multicolumn{5}{c|}{\textbf{Fog}}
  & \multicolumn{5}{c|}{\textbf{Night}}
  & \multicolumn{5}{c|}{\textbf{Rain}}
  & \multicolumn{5}{c}{\textbf{Sun}} \\
\cmidrule(lr){2-6} \cmidrule(lr){7-11} \cmidrule(lr){12-16} \cmidrule(lr){17-21} \cmidrule(lr){22-26}
~ 
  & Ours  & SED   & CAT-Seg & FC-CLIP & EBSeg
  & Ours  & SED   & CAT-Seg & FC-CLIP & EBSeg
  & Ours  & SED   & CAT-Seg & FC-CLIP & EBSeg
  & Ours  & SED   & CAT-Seg & FC-CLIP & EBSeg
  & Ours  & SED   & CAT-Seg & FC-CLIP & EBSeg \\
\midrule
Seen
  & \cellcolor{bestLight}49.14 & 36.72 & \cellcolor{secondLight}43.92 & 28.07 & \cellcolor{thirdLight}38.10
  
  & \cellcolor{bestLight}49.59 & \cellcolor{thirdLight}38.28 & \cellcolor{secondLight}41.43 & 26.06 & 35.76
  
  & \cellcolor{bestLight}41.63 & \cellcolor{secondLight}34.99 & \cellcolor{thirdLight}34.89 & 23.03 & 26.03
  
  & \cellcolor{bestLight}47.15 & \cellcolor{thirdLight}37.35 & \cellcolor{secondLight}41.43 & 25.59 & 31.96
  
  & \cellcolor{bestLight}55.16 & 41.68 & \cellcolor{secondLight}45.10 & 26.52 & \cellcolor{thirdLight}42.63 \\

Unseen
  & \cellcolor{bestLight}10.96 & \cellcolor{secondLight}8.74 & 3.69 & \cellcolor{thirdLight}4.05 & 2.92
  & \cellcolor{bestLight}12.15 & \cellcolor{secondLight}8.40 & 3.54 & 4.32 & \cellcolor{thirdLight}4.48
  & \cellcolor{bestLight}9.52 & \cellcolor{secondLight}7.27 & 2.30 & \cellcolor{thirdLight}4.09 & 2.09
  & \cellcolor{bestLight}12.27 & \cellcolor{secondLight}8.95 & \cellcolor{thirdLight}5.81 & 4.60 & 5.57
  & \cellcolor{bestLight}11.46 & \cellcolor{secondLight}8.12 & 4.91 & 4.73 & \cellcolor{thirdLight}6.14 \\

\midrule
Mean & \cellcolor{bestLight}33.89 & \cellcolor{thirdLight}24.40 & \cellcolor{secondLight}26.22 & 17.50 & 22.62 

     & \cellcolor{bestLight}35.80 & \cellcolor{secondLight}25.25 & \cellcolor{thirdLight}24.80 & 17.26 & 22.00 
     
     & \cellcolor{bestLight}29.89 & \cellcolor{secondLight}22.79 & \cellcolor{thirdLight}20.56 & 14.93 & 15.50 
     
     & \cellcolor{bestLight}33.46 & \cellcolor{thirdLight}25.18 & \cellcolor{secondLight}26.53 & 16.59 & 20.35 
     
     & \cellcolor{bestLight}38.49 & \cellcolor{thirdLight}27.14 & \cellcolor{secondLight}28.21 & 16.93 & 26.41 \\

\bottomrule
\end{tabular}%
}
\label{tab:seen}
\end{table}

\subsection{Ablation Study}

\begin{table*}[!t]
  \centering
  \setlength{\abovecaptionskip}{0.1cm}
  \setlength{\belowcaptionskip}{-0.3cm}
  \caption{Performance comparison of the proposed \textbf{Vireo} against other DGSS methods under the \textit{GTA5\,$\rightarrow$\,Citys.+BDD.+Map.} setting.}
  \label{tab:VFM+}
  \vspace{1em}

  \begin{minipage}[t]{0.49\linewidth}
    \centering
    \caption*{EVA02 (Large) \cite{EVA,EVA02}}
    \renewcommand\arraystretch{1.1}
    \setlength\tabcolsep{3pt}
    \resizebox{\linewidth}{!}{%
      \begin{tabular}{l|c|ccc|c}
        \hline
        
        \textbf{Fine-tune Method} &
        \multicolumn{1}{c|}{\begin{tabular}[c]{@{}c@{}}Trainable\\Params$^*$\end{tabular}} &
        \multicolumn{3}{c|}{\textbf{mIoU}} & \textbf{Avg.} \\
        \cline{3-5}
        & & Citys. & BDD. & Map. & \\ \hline
        Full                       & 304.24M & 62.1 & 56.2 & 64.6 & 60.9\cellcolor[HTML]{EFEFEF}\\
        +AdvStyle\cite{AdvStyle}   & 304.24M & 63.1 & 56.4 & 64.0 & 61.2\cellcolor[HTML]{EFEFEF}\\
        +PASTA\cite{PASTA}         & 304.24M & 61.8 & 57.1 & 63.6 & 60.8\cellcolor[HTML]{EFEFEF}\\
        +GTR-LTR\cite{gtrltr}      & 304.24M & 59.8 & 57.4 & 63.2 & 60.1\cellcolor[HTML]{EFEFEF}\\ \hline
        Freeze                     & ~~0.00M & 56.5 & 53.6 & 58.6 & 56.2\cellcolor[HTML]{EFEFEF}\\
        +AdvStyle\cite{AdvStyle}   & ~~0.00M & 51.4 & 51.6 & 56.5 & 53.2\cellcolor[HTML]{EFEFEF}\\
        +PASTA\cite{PASTA}         & ~~0.00M & 57.8 & 52.3 & 58.5 & 56.2\cellcolor[HTML]{EFEFEF}\\
        +GTR-LTR\cite{gtrltr}      & ~~0.00M & 52.5 & 52.8 & 57.1 & 54.1\cellcolor[HTML]{EFEFEF}\\
        +LoRA\cite{lora}           & ~~1.18M & 55.5 & 52.7 & 58.3 & 55.5\cellcolor[HTML]{EFEFEF}\\
        +AdaptFormer\cite{adaptformer}
                                   & ~~3.17M & 63.7 & 59.9 & 64.2 & 62.6\cellcolor[HTML]{EFEFEF}\\
        +VPT\cite{vpt}             & ~~3.69M & 62.2 & 57.7 & 62.5 & 60.8\cellcolor[HTML]{EFEFEF}\\
        +REIN\cite{REIN}           & ~~2.99M & 65.3\cellcolor{thirdLight}  & 60.5\cellcolor{thirdLight} & 64.9\cellcolor{thirdLight} & 63.6\cellcolor{thirdLight}\\
        +FADA\cite{FADA}           & 11.65M  & 66.7\cellcolor{secondLight} & 61.9\cellcolor{secondLight} & 66.1\cellcolor{secondLight} & 64.9\cellcolor{secondLight}\\
        \textbf{+Vireo (Ours)}     & ~~3.78M & \textbf{68.5}\cellcolor{bestLight} & \textbf{62.1}\cellcolor{bestLight} & \textbf{67.4}\cellcolor{bestLight} & \textbf{66.0}\cellcolor{bestLight}\\
        \hline
      \end{tabular}}
    \vspace{0.4em}
    
  \end{minipage}
  \hfill
  \begin{minipage}[t]{0.49\linewidth}
    \centering
    \caption*{DINOv2 (Large) \cite{DINOV2}}
    \renewcommand\arraystretch{1.1}
    \setlength\tabcolsep{3pt}
    \resizebox{\linewidth}{!}{%
      \begin{tabular}{l|c|ccc|c}
        \hline
        \textbf{Fine-tune Method} &
        \multicolumn{1}{c|}{\begin{tabular}[c]{@{}c@{}}Trainable\\Params$^*$\end{tabular}} &
        \multicolumn{3}{c|}{\textbf{mIoU}} & \textbf{Avg.} \\
        \cline{3-5}
        & & Citys. & BDD. & Map. & \\ \hline
        Full                       & 304.20M & 63.7 & 57.4 & 64.2 & 61.7\cellcolor[HTML]{EFEFEF}\\
        +AdvStyle\cite{AdvStyle}   & 304.20M & 60.8 & 58.0 & 62.5 & 60.4\cellcolor[HTML]{EFEFEF}\\
        +PASTA\cite{PASTA}         & 304.20M & 62.5 & 57.2 & 64.7 & 61.5\cellcolor[HTML]{EFEFEF}\\
        +GTR-LTR\cite{gtrltr}      & 304.20M & 62.7 & 57.4 & 64.5 & 61.6\cellcolor[HTML]{EFEFEF}\\ \hline
        Freeze                     & ~~0.00M & 63.3 & 56.1 & 63.9 & 61.1\cellcolor[HTML]{EFEFEF}\\
        +AdvStyle\cite{AdvStyle}   & ~~0.00M & 61.5 & 55.1 & 63.9 & 60.1\cellcolor[HTML]{EFEFEF}\\
        +PASTA\cite{PASTA}         & ~~0.00M & 62.1 & 57.2 & 64.5 & 61.3\cellcolor[HTML]{EFEFEF}\\
        +GTR-LTR\cite{gtrltr}      & ~~0.00M & 60.2 & 57.7 & 62.2 & 60.0\cellcolor[HTML]{EFEFEF}\\
        +LoRA\cite{lora}           & ~~0.79M & 65.2 & 58.3 & 64.6 & 62.7\cellcolor[HTML]{EFEFEF}\\
        +AdaptFormer\cite{adaptformer}
                                   & ~~3.17M & 64.9 & 59.0 & 64.2 & 62.7\cellcolor[HTML]{EFEFEF}\\
        +VPT\cite{vpt}             & ~~3.69M & 65.2 & 59.4 & 65.5 & 63.3\cellcolor[HTML]{EFEFEF}\\
        +REIN\cite{REIN}           & ~~2.99M & 66.4\cellcolor{thirdLight}  & 60.4\cellcolor{thirdLight} & 66.1\cellcolor{thirdLight} & 64.3\cellcolor{thirdLight}\\
        +FADA\cite{FADA}           & 11.65M  & 68.2\cellcolor{secondLight} & 62.0\cellcolor{secondLight} & 68.1\cellcolor{secondLight} & 66.1\cellcolor{secondLight}\\
        \textbf{+Vireo (Ours)}     & ~~3.78M & \textbf{70.7}\cellcolor{bestLight} & \textbf{62.9}\cellcolor{bestLight} & \textbf{69.6}\cellcolor{bestLight} & \textbf{67.7}\cellcolor{bestLight}\\
        \hline
      \end{tabular}}
    \vspace{0.4em}
  \end{minipage}

\end{table*}

\begin{figure}[!t]
    \centering
    \includegraphics[width=1\linewidth]{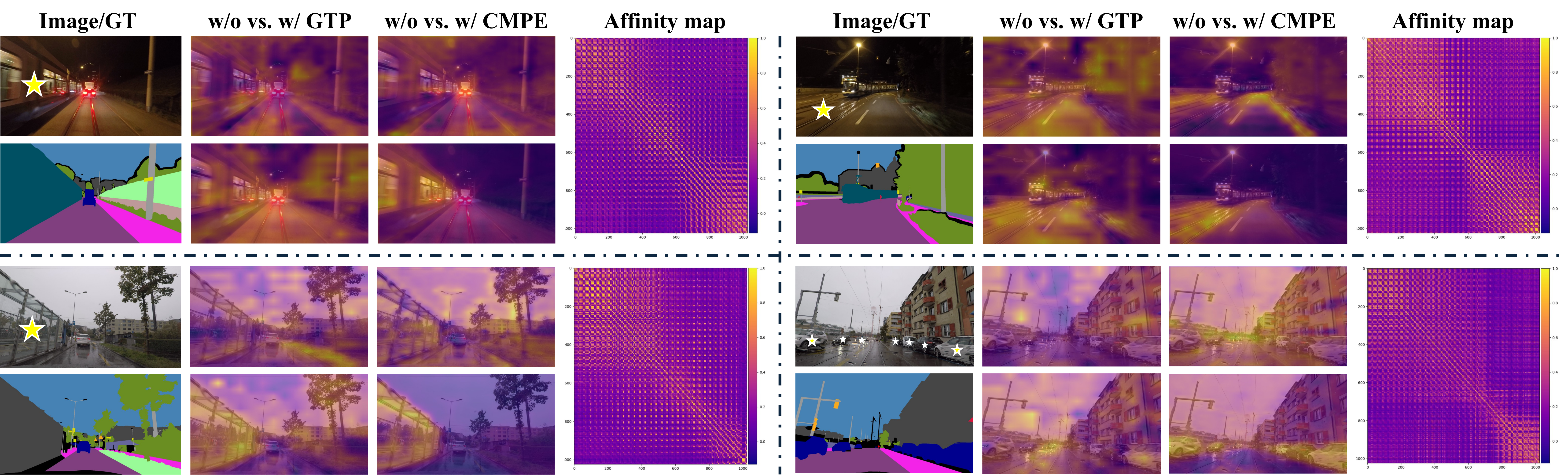}
    \caption{Visualizations of attention maps and affinity maps under differnet scenes, where CMPE denotes the Coarse Mask Prior Embedding, GTQ represents the GeoText Query, respectively.}
    \label{fig:attention}
\end{figure}

\begin{figure}[t]
  \centering
  \begin{minipage}[t]{0.48\columnwidth}\vspace{0pt}
    \centering
    \includegraphics[width=\linewidth]{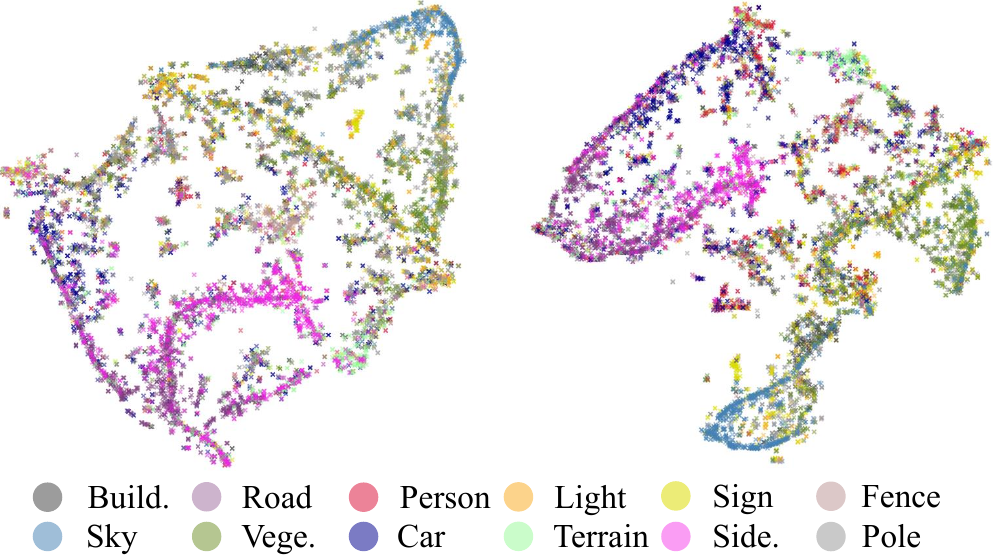}
    \captionof{figure}{t-SNE embeddings of the feature space. \textbf{Left}: original source-domain dataset. \textbf{Right}: our \textbf{Vireo} after adaptation on \textit{Cityscapes $\rightarrow$ ACDC + BDD 100k + Mapillary}. Each point is coloured by its semantic class.}
    \label{fig:tsne}
  \end{minipage}\hfill
  \begin{minipage}[t]{0.48\columnwidth}\vspace{0pt}
    \centering
    \scriptsize
    \setlength{\abovecaptionskip}{0.1cm}
    \setlength{\belowcaptionskip}{-0.3cm}
    \renewcommand\arraystretch{1.0}
    \setlength\tabcolsep{3.0pt}
    \resizebox{\textwidth}{!}{%
    \begin{tabular}{l|cccccc}
      \hline
      Configurations & Snow & Night & Fog & Rain & BDD. & Map. \\
      \hline
      REIN~\cite{REIN}          & 70.6 & 55.9 & 79.5 & 72.5 & 63.5 & 74.0 \\
      $+$ concat $f^d_i$         & 70.8 & 56.1 & 79.4 & 72.8 & 63.6 & 74.2 \\
      $+$ Prompt Depth Anything  & 70.4 & 55.5 & 79.7 & 72.0 & 63.8 & 73.9 \\
      $+$ Depth Anything V2      & 71.5 & 56.7 & 80.5 & 73.3 & 64.4 & 74.5 \\
      $+$ DA + AO               & 72.2\cellcolor{thirdLight}
                                & 57.4\cellcolor{thirdLight}
                                & 80.9\cellcolor{thirdLight}
                                & 74.2\cellcolor{thirdLight}
                                & 65.1\cellcolor{thirdLight}
                                & 75.0\cellcolor{thirdLight} \\
      $+$ DOV-VEH               & 70.9 & 56.2 & 79.8 & 72.8 & 63.7 & 74.2 \\
      $+$ CMPE                  & 71.6 & 56.9 & 80.2 & 73.4 & 64.1 & 74.6 \\
      $+$ GeoText Query       & 74.0\cellcolor{secondLight}
                                & 58.4\cellcolor{secondLight}
                                & 81.1\cellcolor{secondLight}
                                & 74.8\cellcolor{secondLight}
                                & 65.3\cellcolor{secondLight}
                                & 75.3\cellcolor{secondLight} \\
      \textbf{Vireo}            & 76.2\cellcolor{bestLight}
                                & 60.6\cellcolor{bestLight}
                                & 82.3\cellcolor{bestLight}
                                & 76.3\cellcolor{bestLight}
                                & 66.7\cellcolor{bestLight}
                                & 76.0\cellcolor{bestLight} \\
      \hline
    \end{tabular}
    }
    \captionof{table}{Ablation studies on component configurations of the proposed \textbf{Vireo} under the \textit{Citys.\,$\rightarrow$\,ACDC with Snow, Night, Fog, Rain + BDD. + Map.} generalization setting. Top three results are highlighted as \colorbox{bestLight}{\textbf{best}}, \colorbox{secondLight}{second}, and \colorbox{thirdLight}{third} (\%).}
    \label{tab:ablation}
  \end{minipage}
\end{figure}

\begin{table*}[tbp]
  \centering
  \setlength{\abovecaptionskip}{0.1cm}
  \setlength{\belowcaptionskip}{-0.3cm}

  \begin{minipage}[t]{0.48\linewidth}
    \centering
    \caption*{\textbf{CLIP} (ViT-L) \cite{CLIP}}
    \renewcommand\arraystretch{1.1}
    \setlength\tabcolsep{3pt}
    \resizebox{\linewidth}{!}{%
    \begin{tabular}{l|c|ccc|c}
      \hline
      \multirow{2}{*}{Fine-tune} &
      \multirow{2}{*}{\begin{tabular}[c]{@{}c@{}}Trainable\\Params$^*$\end{tabular}} &
      \multicolumn{3}{c|}{mIoU} & \multirow{2}{*}{Avg.} \\
      \cline{3-5}
      & & Citys & BDD & Map & \\ \hline
      Full   & 304.15M & 51.3 & 47.6 & 54.3 & 51.1\cellcolor[HTML]{EFEFEF} \\
      Freeze & ~~0.00M & 53.7 & 48.7 & 55.0 & 52.4\cellcolor[HTML]{EFEFEF} \\
      REIN   & ~~2.99M & 57.1\cellcolor{thirdLight} & 54.7\cellcolor{thirdLight} & 60.5\cellcolor{thirdLight} & 57.4\cellcolor{thirdLight} \\
      FADA   & 11.65M  & 58.7\cellcolor{secondLight} & 55.8\cellcolor{secondLight} & 62.1\cellcolor{secondLight} & 58.9\cellcolor{secondLight} \\
      \textbf{Vireo (Ours)} & ~~3.78M & \textbf{60.5}\cellcolor{bestLight} & \textbf{57.5}\cellcolor{bestLight} & \textbf{64.1}\cellcolor{bestLight} & \textbf{60.7}\cellcolor{bestLight} \\

      \hline
    \end{tabular}}
  \end{minipage}
  \hfill
  \begin{minipage}[t]{0.48\linewidth}
    \centering
    \caption*{\textbf{SAM} (Huge) \cite{SAM}}
    \renewcommand\arraystretch{1.1}
    \setlength\tabcolsep{3pt}
    \resizebox{\linewidth}{!}{%
    \begin{tabular}{l|c|ccc|c}
      \hline
      \multirow{2}{*}{Fine-tune} &
      \multirow{2}{*}{\begin{tabular}[c]{@{}c@{}}Trainable\\Params$^*$\end{tabular}} &
      \multicolumn{3}{c|}{mIoU} & \multirow{2}{*}{Avg.} \\
      \cline{3-5}
      & & Citys & BDD & Map & \\ \hline
      Full   & 632.18M & 57.6 & 51.7 & 61.5 & 56.9\cellcolor[HTML]{EFEFEF} \\
      Freeze & ~~0.00M & 57.0 & 47.1 & 58.4 & 54.2\cellcolor[HTML]{EFEFEF} \\
      REIN   & ~~4.51M & 59.6\cellcolor{thirdLight} & 52.0\cellcolor{thirdLight} & 62.1\cellcolor{thirdLight} & 57.9\cellcolor{thirdLight} \\
      FADA   & 16.59M  & 61.0\cellcolor{secondLight} & 53.2\cellcolor{secondLight} & 63.4\cellcolor{secondLight} & 60.0\cellcolor{secondLight} \\
      \textbf{Vireo (Ours)} & ~~5.30M & \textbf{64.5}\cellcolor{bestLight} & \textbf{59.0}\cellcolor{bestLight} & \textbf{66.0}\cellcolor{bestLight} & \textbf{63.2}\cellcolor{bestLight} \\
      \hline
    \end{tabular}}
  \end{minipage}

  \vspace{1em} 

  \begin{minipage}[t]{0.48\linewidth}
    \centering
    \caption*{\textbf{EVA02} (Large) \cite{EVA,EVA02}}
    \renewcommand\arraystretch{1.1}
    \setlength\tabcolsep{3pt}
    \resizebox{\linewidth}{!}{%
    \begin{tabular}{l|c|ccc|c}
      \hline
      \multirow{2}{*}{Fine-tune} &
      \multirow{2}{*}{\begin{tabular}[c]{@{}c@{}}Trainable\\Params$^*$\end{tabular}} &
      \multicolumn{3}{c|}{mIoU} & \multirow{2}{*}{Avg.} \\
      \cline{3-5}
      & & Citys & BDD & Map & \\ \hline
      Full   & 304.24M & 62.1 & 56.2 & 64.6 & 60.9\cellcolor[HTML]{EFEFEF} \\
      Freeze & ~~0.00M & 56.5 & 53.6 & 58.6 & 56.2\cellcolor[HTML]{EFEFEF} \\
      REIN   & ~~2.99M & 65.3\cellcolor{thirdLight} & 60.5\cellcolor{thirdLight} & 64.9\cellcolor{thirdLight} & 63.6\cellcolor{thirdLight} \\
      FADA   & 11.65M  & 66.7\cellcolor{secondLight} & \textbf{61.9}\cellcolor{secondLight} & 66.1\cellcolor{secondLight} & 64.9\cellcolor{secondLight} \\
      \textbf{Vireo (Ours)} & ~~3.78M & \textbf{68.5}\cellcolor{bestLight} & \textbf{62.1}\cellcolor{bestLight} & \textbf{67.4}\cellcolor{bestLight} & \textbf{66.0}\cellcolor{bestLight}\\
      \hline
    \end{tabular}}
  \end{minipage}
  \hfill
  \begin{minipage}[t]{0.48\linewidth}
    \centering
    \caption*{\textbf{DINOv2} (Large) \cite{DINOV2}}
    \renewcommand\arraystretch{1.1}
    \setlength\tabcolsep{3pt}
    \resizebox{\linewidth}{!}{%
    \begin{tabular}{l|c|ccc|c}
      \hline
      \multirow{2}{*}{Fine-tune} &
      \multirow{2}{*}{\begin{tabular}[c]{@{}c@{}}Trainable\\Params$^*$\end{tabular}} &
      \multicolumn{3}{c|}{mIoU} & \multirow{2}{*}{Avg.} \\
      \cline{3-5}
      & & Citys & BDD & Map & \\ \hline
      Full   & 304.20M & 63.7 & 57.4 & 64.2 & 61.7\cellcolor[HTML]{EFEFEF} \\
      Freeze & ~~0.00M & 63.3 & 56.1 & 63.9 & 61.1\cellcolor[HTML]{EFEFEF} \\
      REIN   & ~~2.99M & 66.4\cellcolor{thirdLight} & 60.4\cellcolor{thirdLight} & 66.1\cellcolor{thirdLight} & 64.3\cellcolor{thirdLight} \\
      FADA   & 11.65M  & 68.2\cellcolor{secondLight} & 62.0\cellcolor{secondLight} & 68.1\cellcolor{secondLight} & 66.1\cellcolor{secondLight} \\
      \textbf{Vireo (Ours)} & ~~3.78M & \textbf{70.7}\cellcolor{bestLight} & \textbf{62.9}\cellcolor{bestLight} & \textbf{69.6}\cellcolor{bestLight} & \textbf{67.7}\cellcolor{bestLight}\\
      \hline
    \end{tabular}}
  \end{minipage}

  \caption{Performance comparison of \textbf{Vireo} across multiple VFMs under the \textit{GTA5 $\rightarrow$ Citys.\,+\,BDD.\,+\,Map.} setting. $^*$Trainable parameters in backbone. Top three results are \colorbox{bestLight}{best}, \colorbox{secondLight}{second}, \colorbox{thirdLight}{third}. (\%)}
  \label{tab:VFMs}
\end{table*}

\textbf{Robust Performance Gains.} In Table~\ref{tab:VFM+}, we compare parameter overhead and mIoU of lightweight tuning methods on EVA02-Large and DINOv2-Large under the GTA5$\rightarrow$Cityscapes + BDD + Mapillary transfer. With only $\approx3.8$\,M trainable parameters, Vireo leads both: $66.0\%$ on EVA02 (+$1.1\%$ over FADA) and $67.7\%$ on DINOv2 (+$1.6\%$). Other schemes (REIN, LoRA, VPT, AdaptFormer) improve a frozen backbone but fall short of Vireo’s accuracy-efficiency balance. Versus full fine-tuning, Vireo slashes training cost and further improves mIoU, showing its depth-geometry queries generalize across VFMs.

\begin{wrapfigure}[17]{r}{0.25\textwidth} 
  \vspace{-4pt} 
  \centering
  \includegraphics[width=0.25\textwidth]{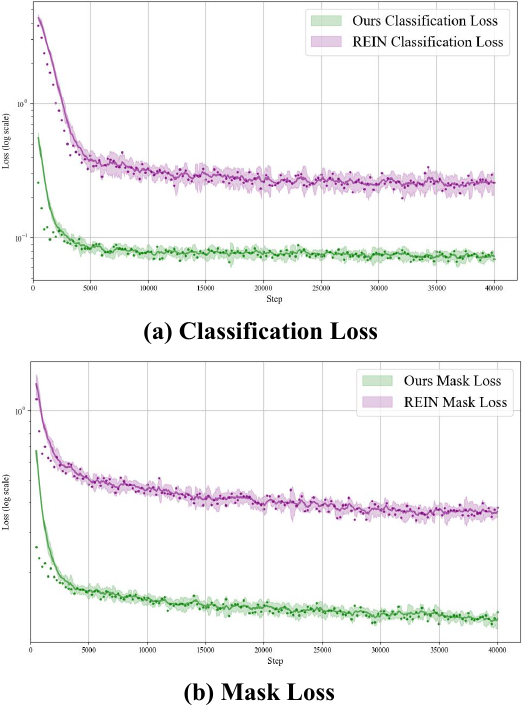}
  \caption{Comparison of Train Loss for Baseline and CMPE Models.}
  \label{fig:loss}      
  \vspace{-4pt}         
\end{wrapfigure}

Table \ref{tab:VFMs} reports the parameter overhead and mIoU of Vireo and several lightweight fine-tuning schemes on four backbones (CLIP-L, SAM-H, EVA02-L, and DINOv2-L). Compared with the prompt-based REIN baseline, Vireo adds only about 0.79 M extra parameters yet achieves the highest average mIoU on every backbone: the gain is most pronounced on the parameter-constrained CLIP-L, and Vireo still surpasses heavier adapters such as FADA by 1–2 mIoU on the larger EVA02-L and DINOv2-L models, underscoring its parameter efficiency and scalability across backbones.

In Table~\ref{tab:ablation}, using Depth Anything V2 alone yields an approximate $0.6\%$ mIoU gain across all six scenarios, and adding depth augmentation with attention optimization (DA + AO) delivers a further $\approx1\%$ improvement. The DOV-VEH and CMPE modules each add $0.5\%$--$0.8\%$, validating mask vectors and dense gradient embedding. GeoText Query on its own provides a substantial $\approx4.4\%$ boost, underscoring the complementary benefits of fusing semantic and geometric cues.

\textbf{Pronounced Attention Focus.} Figure~\ref{fig:attention} further confirm that GeoText Query steer the model toward geometry-sensitive regions, while CMPE strengthens gradients on foreground masks; compared with baselines, Vireo’s focus is tighter on scene structure and semantic boundaries, explaining its consistent advantage in cross-domain and open-vocabulary settings.

\textbf{t-SNE Visualization of DGSS Features.} The feature distributions of the original dataset and our method are visualized in Figure~\ref{fig:tsne}, revealing the superiority of our learned features in forming well-separated semantic clusters. This demonstrates the effectiveness of our domain-generalized visual-textual alignment in structuring the open-vocabulary semantic space.

\section{Conclusion}
This study introduces Vireo, the first single-stage framework that unifies open-vocabulary recognition and domain-generalised semantic segmentation. By integrating frozen visual foundation models, depth-aware geometry and three core modules—GeoText Query, Coarse Mask Prior Embedding and the DOV Vector Embedding Head—it converges faster, concentrates attention on scene structure and outperforms state-of-the-art methods across multiple benchmarks, demonstrating the power of combining textual cues with geometric priors for robust pixel-level perception.

\textbf{Limitations} Vireo assumes a reliable RGB camera, but in rare cases (e.g., occlusion, glare, hardware failure), the stream can be lost, impairing perception. Future work will explore multi‐source setups—automatically switching to lidar, radar, or event‐based cameras when RGB fails—to keep segmentation robust in all conditions.
\FloatBarrier 

\section*{Acknowledgements}
This work was supported in part by the Natural Science Foundation of Xiamen, China, under Grants 3502Z202373036 and 3502Z202371019; in part by the National Natural Science Foundation of China under Grant 42371457; in part by the Natural Science Foundation of Fujian Province, China, under Grants 2025J01345 and 2023J01803; by the 2025 University of Glasgow Early Career Reward for Excellence on ``\emph{Developing Vision–Language Models Enhanced by Geospatial Intelligence}'' and the 2025 University of Glasgow Early Career Mobility Scheme (ECMS) Award; and by the Program of China Scholarship Council under Grant No. 202506380088.

\begingroup
  \small
  \bibliographystyle{IEEEtran}
  \bibliography{ref}
\endgroup

\newpage
\section*{NeurIPS Paper Checklist}

\begin{enumerate}

\item {\bf Claims}
    \item[] Question: Do the main claims made in the abstract and introduction accurately reflect the paper's contributions and scope?
    \item[] Answer: \answerYes{} 
    \item[] Justification: The abstract’s and introduction’s key claims map exactly onto our contributions and scope
    \item[] Guidelines:
    \begin{itemize}
        \item The answer NA means that the abstract and introduction do not include the claims made in the paper.
        \item The abstract and/or introduction should clearly state the claims made, including the contributions made in the paper and important assumptions and limitations. A No or NA answer to this question will not be perceived well by the reviewers. 
        \item The claims made should match theoretical and experimental results, and reflect how much the results can be expected to generalize to other settings. 
        \item It is fine to include aspirational goals as motivation as long as it is clear that these goals are not attained by the paper. 
    \end{itemize}

\item {\bf Limitations}
    \item[] Question: Does the paper discuss the limitations of the work performed by the authors?
    \item[] Answer: \answerYes{} 
    \item[] Justification: In the conclusion section of the paper we discuss the limitations of the paper.
    \item[] Guidelines:
    \begin{itemize}
        \item The answer NA means that the paper has no limitation while the answer No means that the paper has limitations, but those are not discussed in the paper. 
        \item The authors are encouraged to create a separate "Limitations" section in their paper.
        \item The paper should point out any strong assumptions and how robust the results are to violations of these assumptions (e.g., independence assumptions, noiseless settings, model well-specification, asymptotic approximations only holding locally). The authors should reflect on how these assumptions might be violated in practice and what the implications would be.
        \item The authors should reflect on the scope of the claims made, e.g., if the approach was only tested on a few datasets or with a few runs. In general, empirical results often depend on implicit assumptions, which should be articulated.
        \item The authors should reflect on the factors that influence the performance of the approach. For example, a facial recognition algorithm may perform poorly when image resolution is low or images are taken in low lighting. Or a speech-to-text system might not be used reliably to provide closed captions for online lectures because it fails to handle technical jargon.
        \item The authors should discuss the computational efficiency of the proposed algorithms and how they scale with dataset size.
        \item If applicable, the authors should discuss possible limitations of their approach to address problems of privacy and fairness.
        \item While the authors might fear that complete honesty about limitations might be used by reviewers as grounds for rejection, a worse outcome might be that reviewers discover limitations that aren't acknowledged in the paper. The authors should use their best judgment and recognize that individual actions in favor of transparency play an important role in developing norms that preserve the integrity of the community. Reviewers will be specifically instructed to not penalize honesty concerning limitations.
    \end{itemize}

\item {\bf Theory assumptions and proofs}
    \item[] Question: For each theoretical result, does the paper provide the full set of assumptions and a complete (and correct) proof?
    \item[] Answer: \answerYes{} 
    \item[] Justification: The theory assumptions are in the methodology section.
    \item[] Guidelines:
    \begin{itemize}
        \item The answer NA means that the paper does not include theoretical results. 
        \item All the theorems, formulas, and proofs in the paper should be numbered and cross-referenced.
        \item All assumptions should be clearly stated or referenced in the statement of any theorems.
        \item The proofs can either appear in the main paper or the supplemental material, but if they appear in the supplemental material, the authors are encouraged to provide a short proof sketch to provide intuition. 
        \item Inversely, any informal proof provided in the core of the paper should be complemented by formal proofs provided in appendix or supplemental material.
        \item Theorems and Lemmas that the proof relies upon should be properly referenced. 
    \end{itemize}

    \item {\bf Experimental result reproducibility}
    \item[] Question: Does the paper fully disclose all the information needed to reproduce the main experimental results of the paper to the extent that it affects the main claims and/or conclusions of the paper (regardless of whether the code and data are provided or not)?
    \item[] Answer: \answerYes{} 
    \item[] Justification: The model realization and implementation details are provided in the experimental section.
    \item[] Guidelines:
    \begin{itemize}
        \item The answer NA means that the paper does not include experiments.
        \item If the paper includes experiments, a No answer to this question will not be perceived well by the reviewers: Making the paper reproducible is important, regardless of whether the code and data are provided or not.
        \item If the contribution is a dataset and/or model, the authors should describe the steps taken to make their results reproducible or verifiable. 
        \item Depending on the contribution, reproducibility can be accomplished in various ways. For example, if the contribution is a novel architecture, describing the architecture fully might suffice, or if the contribution is a specific model and empirical evaluation, it may be necessary to either make it possible for others to replicate the model with the same dataset, or provide access to the model. In general. releasing code and data is often one good way to accomplish this, but reproducibility can also be provided via detailed instructions for how to replicate the results, access to a hosted model (e.g., in the case of a large language model), releasing of a model checkpoint, or other means that are appropriate to the research performed.
        \item While NeurIPS does not require releasing code, the conference does require all submissions to provide some reasonable avenue for reproducibility, which may depend on the nature of the contribution. For example
        \begin{enumerate}
            \item If the contribution is primarily a new algorithm, the paper should make it clear how to reproduce that algorithm.
            \item If the contribution is primarily a new model architecture, the paper should describe the architecture clearly and fully.
            \item If the contribution is a new model (e.g., a large language model), then there should either be a way to access this model for reproducing the results or a way to reproduce the model (e.g., with an open-source dataset or instructions for how to construct the dataset).
            \item We recognize that reproducibility may be tricky in some cases, in which case authors are welcome to describe the particular way they provide for reproducibility. In the case of closed-source models, it may be that access to the model is limited in some way (e.g., to registered users), but it should be possible for other researchers to have some path to reproducing or verifying the results.
        \end{enumerate}
    \end{itemize}

\item {\bf Open access to data and code}
    \item[] Question: Does the paper provide open access to the data and code, with sufficient instructions to faithfully reproduce the main experimental results, as described in supplemental material?
    \item[] Answer: \answerYes{} 
    \item[] Justification: The dataset is publicly available and the code has been open sourced.
    \item[] Guidelines:
    \begin{itemize}
        \item The answer NA means that paper does not include experiments requiring code.
        \item Please see the NeurIPS code and data submission guidelines (\url{https://nips.cc/public/guides/CodeSubmissionPolicy}) for more details.
        \item While we encourage the release of code and data, we understand that this might not be possible, so “No” is an acceptable answer. Papers cannot be rejected simply for not including code, unless this is central to the contribution (e.g., for a new open-source benchmark).
        \item The instructions should contain the exact command and environment needed to run to reproduce the results. See the NeurIPS code and data submission guidelines (\url{https://nips.cc/public/guides/CodeSubmissionPolicy}) for more details.
        \item The authors should provide instructions on data access and preparation, including how to access the raw data, preprocessed data, intermediate data, and generated data, etc.
        \item The authors should provide scripts to reproduce all experimental results for the new proposed method and baselines. If only a subset of experiments are reproducible, they should state which ones are omitted from the script and why.
        \item At submission time, to preserve anonymity, the authors should release anonymized versions (if applicable).
        \item Providing as much information as possible in supplemental material (appended to the paper) is recommended, but including URLs to data and code is permitted.
    \end{itemize}

\item {\bf Experimental setting/details}
    \item[] Question: Does the paper specify all the training and test details (e.g., data splits, hyperparameters, how they were chosen, type of optimizer, etc.) necessary to understand the results?
    \item[] Answer: \answerYes{} 
    \item[] Justification: The Deployment Details \& Parameter Settings are provided in the experiment section.
    \item[] Guidelines:
    \begin{itemize}
        \item The answer NA means that the paper does not include experiments.
        \item The experimental setting should be presented in the core of the paper to a level of detail that is necessary to appreciate the results and make sense of them.
        \item The full details can be provided either with the code, in appendix, or as supplemental material.
    \end{itemize}

\item {\bf Experiment statistical significance}
    \item[] Question: Does the paper report error bars suitably and correctly defined or other appropriate information about the statistical significance of the experiments?
    \item[] Answer: \answerYes{} 
    \item[] Justification: The evaluation protocols of these segmentation datasets do not require report
the error bars.
    \item[] Guidelines:
    \begin{itemize}
        \item The answer NA means that the paper does not include experiments.
        \item The authors should answer "Yes" if the results are accompanied by error bars, confidence intervals, or statistical significance tests, at least for the experiments that support the main claims of the paper.
        \item The factors of variability that the error bars are capturing should be clearly stated (for example, train/test split, initialization, random drawing of some parameter, or overall run with given experimental conditions).
        \item The method for calculating the error bars should be explained (closed form formula, call to a library function, bootstrap, etc.)
        \item The assumptions made should be given (e.g., Normally distributed errors).
        \item It should be clear whether the error bar is the standard deviation or the standard error of the mean.
        \item It is OK to report 1-sigma error bars, but one should state it. The authors should preferably report a 2-sigma error bar than state that they have a 96\% CI, if the hypothesis of Normality of errors is not verified.
        \item For asymmetric distributions, the authors should be careful not to show in tables or figures symmetric error bars that would yield results that are out of range (e.g. negative error rates).
        \item If error bars are reported in tables or plots, The authors should explain in the text how they were calculated and reference the corresponding figures or tables in the text.
    \end{itemize}

\item {\bf Experiments compute resources}
    \item[] Question: For each experiment, does the paper provide sufficient information on the computer resources (type of compute workers, memory, time of execution) needed to reproduce the experiments?
    \item[] Answer: \answerYes{} 
    \item[] Justification: The computation resources and details are discussed in experiment section.
    \item[] Guidelines:
    \begin{itemize}
        \item The answer NA means that the paper does not include experiments.
        \item The paper should indicate the type of compute workers CPU or GPU, internal cluster, or cloud provider, including relevant memory and storage.
        \item The paper should provide the amount of compute required for each of the individual experimental runs as well as estimate the total compute. 
        \item The paper should disclose whether the full research project required more compute than the experiments reported in the paper (e.g., preliminary or failed experiments that didn't make it into the paper). 
    \end{itemize}
    
\item {\bf Code of ethics}
    \item[] Question: Does the research conducted in the paper conform, in every respect, with the NeurIPS Code of Ethics \url{https://neurips.cc/public/EthicsGuidelines}?
    \item[] Answer: \answerYes{} 
    \item[] Justification:  The authors have read the code of ethics. The experiments are all on public datasets without obeying the code of ethics.
    \item[] Guidelines:
    \begin{itemize}
        \item The answer NA means that the authors have not reviewed the NeurIPS Code of Ethics.
        \item If the authors answer No, they should explain the special circumstances that require a deviation from the Code of Ethics.
        \item The authors should make sure to preserve anonymity (e.g., if there is a special consideration due to laws or regulations in their jurisdiction).
    \end{itemize}

\item {\bf Broader impacts}
    \item[] Question: Does the paper discuss both potential positive societal impacts and negative societal impacts of the work performed?
    \item[] Answer: \answerNA{} 
    \item[] Justification: At the end of the conclusion section, the broader impacts have been discussed.
    \item[] Guidelines:
    \begin{itemize}
        \item The answer NA means that there is no societal impact of the work performed.
        \item If the authors answer NA or No, they should explain why their work has no societal impact or why the paper does not address societal impact.
        \item Examples of negative societal impacts include potential malicious or unintended uses (e.g., disinformation, generating fake profiles, surveillance), fairness considerations (e.g., deployment of technologies that could make decisions that unfairly impact specific groups), privacy considerations, and security considerations.
        \item The conference expects that many papers will be foundational research and not tied to particular applications, let alone deployments. However, if there is a direct path to any negative applications, the authors should point it out. For example, it is legitimate to point out that an improvement in the quality of generative models could be used to generate deepfakes for disinformation. On the other hand, it is not needed to point out that a generic algorithm for optimizing neural networks could enable people to train models that generate Deepfakes faster.
        \item The authors should consider possible harms that could arise when the technology is being used as intended and functioning correctly, harms that could arise when the technology is being used as intended but gives incorrect results, and harms following from (intentional or unintentional) misuse of the technology.
        \item If there are negative societal impacts, the authors could also discuss possible mitigation strategies (e.g., gated release of models, providing defenses in addition to attacks, mechanisms for monitoring misuse, mechanisms to monitor how a system learns from feedback over time, improving the efficiency and accessibility of ML).
    \end{itemize}
    
\item {\bf Safeguards}
    \item[] Question: Does the paper describe safeguards that have been put in place for responsible release of data or models that have a high risk for misuse (e.g., pretrained language models, image generators, or scraped datasets)?
    \item[] Answer: \answerNA{} 
    \item[] Justification: This work does not have such risks.
    \item[] Guidelines:
    \begin{itemize}
        \item The answer NA means that the paper poses no such risks.
        \item Released models that have a high risk for misuse or dual-use should be released with necessary safeguards to allow for controlled use of the model, for example by requiring that users adhere to usage guidelines or restrictions to access the model or implementing safety filters. 
        \item Datasets that have been scraped from the Internet could pose safety risks. The authors should describe how they avoided releasing unsafe images.
        \item We recognize that providing effective safeguards is challenging, and many papers do not require this, but we encourage authors to take this into account and make a best faith effort.
    \end{itemize}

\item {\bf Licenses for existing assets}
    \item[] Question: Are the creators or original owners of assets (e.g., code, data, models), used in the paper, properly credited and are the license and terms of use explicitly mentioned and properly respected?
    \item[] Answer: \answerYes{} 
    \item[] Justification: All external code, data, and pre-trained models we build on are cited, their original licenses are named, and our use conforms to those terms.
    \item[] Guidelines:
    \begin{itemize}
        \item The answer NA means that the paper does not use existing assets.
        \item The authors should cite the original paper that produced the code package or dataset.
        \item The authors should state which version of the asset is used and, if possible, include a URL.
        \item The name of the license (e.g., CC-BY 4.0) should be included for each asset.
        \item For scraped data from a particular source (e.g., website), the copyright and terms of service of that source should be provided.
        \item If assets are released, the license, copyright information, and terms of use in the package should be provided. For popular datasets, \url{paperswithcode.com/datasets} has curated licenses for some datasets. Their licensing guide can help determine the license of a dataset.
        \item For existing datasets that are re-packaged, both the original license and the license of the derived asset (if it has changed) should be provided.
        \item If this information is not available online, the authors are encouraged to reach out to the asset's creators.
    \end{itemize}

\item {\bf New assets}
    \item[] Question: Are new assets introduced in the paper well documented and is the documentation provided alongside the assets?
    \item[] Answer: \answerYes{} 
    \item[] Justification: We will release our code base with included readme files. 
    \item[] Guidelines:
    \begin{itemize}
        \item The answer NA means that the paper does not release new assets.
        \item Researchers should communicate the details of the dataset/code/model as part of their submissions via structured templates. This includes details about training, license, limitations, etc. 
        \item The paper should discuss whether and how consent was obtained from people whose asset is used.
        \item At submission time, remember to anonymize your assets (if applicable). You can either create an anonymized URL or include an anonymized zip file.
    \end{itemize}

\item {\bf Crowdsourcing and research with human subjects}
    \item[] Question: For crowdsourcing experiments and research with human subjects, does the paper include the full text of instructions given to participants and screenshots, if applicable, as well as details about compensation (if any)? 
    \item[] Answer: \answerNA{} 
    \item[] Justification: This work does not involve crowdsourcing nor research with human subject
    \item[] Guidelines:
    \begin{itemize}
        \item The answer NA means that the paper does not involve crowdsourcing nor research with human subjects.
        \item Including this information in the supplemental material is fine, but if the main contribution of the paper involves human subjects, then as much detail as possible should be included in the main paper. 
        \item According to the NeurIPS Code of Ethics, workers involved in data collection, curation, or other labor should be paid at least the minimum wage in the country of the data collector. 
    \end{itemize}

\item {\bf Institutional review board (IRB) approvals or equivalent for research with human subjects}
    \item[] Question: Does the paper describe potential risks incurred by study participants, whether such risks were disclosed to the subjects, and whether Institutional Review Board (IRB) approvals (or an equivalent approval/review based on the requirements of your country or institution) were obtained?
    \item[] Answer: \answerNA{} 
    \item[] Justification: This work does not involve crowdsourcing nor research with human subjects.
    \item[] Guidelines:
    \begin{itemize}
        \item The answer NA means that the paper does not involve crowdsourcing nor research with human subjects.
        \item Depending on the country in which research is conducted, IRB approval (or equivalent) may be required for any human subjects research. If you obtained IRB approval, you should clearly state this in the paper. 
        \item We recognize that the procedures for this may vary significantly between institutions and locations, and we expect authors to adhere to the NeurIPS Code of Ethics and the guidelines for their institution. 
        \item For initial submissions, do not include any information that would break anonymity (if applicable), such as the institution conducting the review.
    \end{itemize}

\item {\bf Declaration of LLM usage}
    \item[] Question: Does the paper describe the usage of LLMs if it is an important, original, or non-standard component of the core methods in this research? Note that if the LLM is used only for writing, editing, or formatting purposes and does not impact the core methodology, scientific rigorousness, or originality of the research, declaration is not required.
    \item[] Answer: \answerNA{} 
    \item[] Justification: No large language models were used as part of the research methodology
    \item[] Guidelines:
    \begin{itemize}
        \item The answer NA means that the core method development in this research does not involve LLMs as any important, original, or non-standard components.
        \item Please refer to our LLM policy (\url{https://neurips.cc/Conferences/2025/LLM}) for what should or should not be described.
    \end{itemize}

\end{enumerate}

\end{document}